\documentclass{article}

\usepackage{arxiv}

\usepackage[utf8]{inputenc} 
\usepackage[T1]{fontenc}    
\usepackage{hyperref}       
\usepackage{url}            
\usepackage{booktabs}       
\usepackage{amsfonts}       
\usepackage{nicefrac}       
\usepackage{microtype}      
\usepackage{lipsum}
\usepackage{graphicx}
\usepackage{CJKutf8}
\graphicspath{ {./images/} }
\usepackage{amsmath,amsthm,amssymb,amsfonts,amscd,keyval}
\usepackage{bm}
\usepackage{mathtools}
\usepackage{multirow}
\usepackage{dsfont}
\usepackage{float}
\usepackage{xspace}
\usepackage{subfig}
\usepackage{bbding}
\usepackage{makecell}
 \usepackage{threeparttable}
 \usepackage[normalem]{ulem}
\useunder{\uline}{\ul}{}
\title{TSEN: Transformer and Snowball Graph Convolution Learning for Brain Functional Network Classification}

\author{%
Jinlong Hu\thanks{Corresponding author. Email: jlhu@scut.edu.cn} \quad Yangmin Huang \quad Shoubin Dong \quad\\
Guangdong Key Lab of Communication and Computer Network\\
School of Computer Science and Engineering\\
South China University of Technology\\
Guangzhou, China\\
}

\begin{document}
\begin{CJK}{UTF8}{gbsn}
\maketitle
\begin{abstract}
Advanced deep learning methods, especially graph neural networks (GNNs), are increasingly expected to learn from brain functional network data and predict brain disorders. In this paper, we proposed a novel Transformer and snowball encoding networks (TSEN) for brain functional network classification, which introduced Transformer architecture with graph snowball connection into GNNs for learning whole-graph representation. TSEN combined graph snowball connection with graph Transformer by snowball encoding layers, which enhanced the power to capture multi-scale information and global patterns of brain functional networks. TSEN also introduced snowball graph convolution as position embedding in Transformer structure, which was a simple yet effective method for capturing local patterns naturally. We evaluated the proposed model by two large-scale brain functional network datasets from autism spectrum disorder and major depressive disorder respectively, and the results demonstrated that TSEN outperformed the state-of-the-art GNN models and the graph-transformer based GNN models. 
\end{abstract}


\section{Introduction}
A growing number of deep learning methods, especially graph neural networks (GNNs), have been developed for learning on brain functional network data and predicting brain disorders~\cite{bessadok2022graph,PAN2022,WANG2023}. Graph classification is one of the most prominent learning tasks of GNNs, which typically learns a graph-level representation for graph classification in an end-to-end fashion. To learn an effective representation, GNNs were expected to capture the global patterns as well as the local patterns and their interaction~\cite{RN2}. There have been various attempts to enhance the expressive power of GNNs, for example, enlarging the receptive field~\cite{RN5}, using effective graph pooling operations~\cite{RN6,RN7}, and relieving the over-smoothing caused by increased network depth~\cite{RN8,RN18}. Brain functional networks are usually constructed from neuroimaging data such as functional Magnetic Resonance Imaging (fMRI), where the nodes are defined by Regions of Interest (ROIs) based on a brain atlas, and the edges are defined by pairwise correlations between the fMRI signal series of ROIs. Compared with data from other fields, brain functional networks are usually complex and noisy, and the data size is usually limited~\cite{constable2012challenges}. Given brain functional network data, it is difficult to develop efficient graph-level learning approaches to produce compact vector representations which can be optimized for classification task.

Recently, Transformer~\cite{RN9} has been shown excel performance in nature language processing (NLP), such as Bert~\cite{RN10} and GPT~\cite{RN11}, and this architecture was also introduced for graph structure data~\cite{kan2022brain}. Transformer alleviates the limitations related to the sparse message passing mechanism from GNNs, such as over-smoothing and over-squashing~\cite{rampavsek2022recipe}. And Transformer has been shown to be good at capturing the long-range dependency and global information in graph~\cite{RN15}. Some studies tried to properly incorporate structural information of graphs into GNN models via Transformer, such as using graph structure encoding in Graphormer~\cite{RN16}, and subgraph representation in structure-aware transformer (SAT)~\cite{RN17}. However, it is still a challenge to develop targeted Transformer architecture for whole-graph representation learning
in classifying brain functional networks. 

In this paper, we proposed a Transformer and snowball encoding networks (TSEN) for brain functional network classification, which introduced Transformer architecture with snowball connection into GNNs for learning whole-graph representations from given brain networks. TSEN learns multi-scale information of nodes with snowball encoding, and read-out the node information with global attention to obtain graph level representation for classification. A snowball encoding function was proposed to capture short-range and long-range information by combining graph convolution with dense connection structure and graph Transformer. The proposed method was evaluated by two large-scale brain functional network datasets from autism spectrum disorder and major depressive disorder, respectively. The results showed that our proposed method achieved the best performance compared with the state-of-the-art typical GNN models and the graph-transformer based GNN models. Furthermore, we validated the effectiveness of TSEN with ablation experiments, and further analyzed the representation capacity of TSEN and the influence of input graphs on the model.

The main contributions are summarized as follows. 
\begin{itemize}
\item[-]
We proposed a novel graph Transformer architecture with snowball connection for graph neural networks, which enhanced the power to capture multi-scale information and global patterns of brain functional networks. And the proposed model efficiently learned positional and structural information from node features and edge features in the brain functional networks, where node feature was defined by the node’s corresponding row in the brain network adjacency matrix~\cite{kan2022brain, RN39} and edges were binarized by thresholding on elements of the adjacency matrix.
\item[-] We introduced snowball graph convolution as position embedding in Transformer structure, which was a simple yet effective method to capture structure information naturally with Transformer structure from brain functional networks. 
\item[-] We performed extensive ablation studies that evaluated contribution of graph convolution, snowball, feed-forward network, and self-attention components on the benchmarking datasets. And we performed extensive analysis that demonstrated the powerful representation of the proposed model with visualization and similarity comparison. 
\end{itemize}
The rest of this paper is organized as follows. Section~\ref{sec:headings} reviews the related literature. Section~\ref{sec:notation} describes the notation. Section~\ref{sec:method} presents the proposed TSEN. Section~\ref{sec:experiments} reports the experiments and results. Section~\ref{sec:analysis} reports the analysis of TSEN model. Finally, Section~\ref{sec:conclusion} concludes this paper.

\section{Related Work}
\label{sec:headings}
\paragraph{Graph Classification with GNNs.}
Given a brain functional network, graph classification with GNNs typically learns a graph-level representation for graph classification. Graph-level representation learning with GNNs mainly includes graph pooling and coarsening method. The graph pooling methods usually use global pooling methods to read-out the graphs such as SAGPool~\cite{RN6}. The coarsening methods apply grouping nodes into similarity clusters to extract global representations, such as Diff-Pool~\cite{RN7} and Graclus multi-layer clustering~\cite{RN19}.

Recent works showed that more sophisticated layers for aggregating or read-out the graphs, such as clustering layers, may not bring any benefits on many real-world graph data~\cite{RN20}. To capture better structure information for graph classification, GraphSAGE~\cite{RN44} used random edge sampling methods, Graph Isomorphism Networks (GIN)~\cite{RN34} and subgraph isomorphism counting method~\cite{RN23} used graph isomorphism-based methods to achieve the ability as the Weisfeiler-Lehman test.

Residual and dense connections~\cite{RN24,RN25} have demonstrated their great improvement in traditional neural networks such as convolutional neural networks (CNNs), and the structures of residual and dense connection were also applied to GNN domain~\cite{RN8,RN18,RN26,RN27}. Residual connections could mitigate the problem of over-smoothing and improve the performance of GNNs with deeper structure~\cite{RN8,RN26}. Dense connections use the structure of snowball connections and consiTSENtly up to more convolutional layers, and the snowball structure could concatenate multi-scale features incrementally~\cite{RN18}. The multi-scale information of graphs could be helpful for enhancing the expressiveness of GNNs on graph classification~\cite{RN28,RN29,RN45}.
\paragraph{Transformer on Graph Data}
Transformer~\cite{RN9} has been successfully used in sequence modeling such as natural language modeling, since its special design of multi-head attention could capture the long-range dependency and relationship. In the Transformer structure for GNNs, positional embedding methods were used to improve the power of capturing local patterns and structure on graphs. For example, Graph-Bert~\cite{RN30} used three positional embedding methods, Weisfeiler-Lehman absolute role embedding, intimacy-based relative positional embedding, and hop-based relative distance embedding, to extract substructure information. Generalization of Transformer (GT)~\cite{RN31} proposed a positional embedding method, Laplacian eigenvector, to improve its performance. Graphormer~\cite{RN16} was recently proposed to raise positional encoding methods with edge, spatial and centrality, designed for solving large-scale datasets of molecules.

Besides using positional embedding for graphs in Transformer, standard GNN module could be combined with Transformer to extract structure information, and use a “readout” mechanism similar to “cls” in NLP to obtain a global graph embedding~\cite{RN15}. Furthermore, Chen et al.~\cite{RN17} suggested that using GNN-based methods was effective for generating subgraph representation and introducing structure information naturally to Transformer structure. In this paper, we used graph convolution as the position embedding method and introduced the snowball structure into Transformer, to learn whole-graph representations for brain functional network classification.
\section{Notation}
\label{sec:notation}
\paragraph{Graph}
Let ${\rm{\;G}} = \left( {V,E,{\bf{X}}} \right)$ be a graph, where $V = \left\{ {{v_0}, \ldots ,{v_{{\rm{n}} - 1}}} \right\}$ denotes the set of nodes of the graph, $E \subseteq V \times V$ is the set of edges, and ${\bf{X}} \in {{\mathbb R}^{{\rm{n}} \times {\rm{s}}}}$ is a multi-variate signal on the graph nodes, with ${\rm{s}}$ represents the dimension of node features.
\paragraph{Graph convolution operating}
The graph convolutional operator $f\left(  \cdot  \right)$  is defined as follows:
\begin{equation}
{\bf{H}} = f\left( {\bf{X}} \right) = activation\left( {{\bf{LXW}}} \right)
\label{eq:gcn}
\end{equation}
Where ${\bf{H}} \in {{\mathbb R}^{{\rm{n}} \times {\rm{s'}}}}$ denotes the convoluted node features, ${\bf{W}} \in {{\mathbb R}^{{\rm{s}} \times {\rm{s'}}}}$ represents the weighted matrix, with ${\rm{s'}}$ represents the dimension of convoluted node features, $activation\left(  \cdot  \right)$ is the activation function; ${\rm{\;}}{\bf{L}}$ is the normalized graph Laplacian which is defined by: 
\begin{equation}
{\bf{L}} = {\bf{I}} - {{\bf{D}}^{ - \left( {\frac{1}{2}} \right)}}{\bf{A}}{{\bf{D}}^{ - \left( {\frac{1}{2}} \right)}}
\label{eq:laplacian}
\end{equation}
Where ${\bf{A}} \in {{\mathbb R}^{{\rm{n}} \times {\rm{n}}}}$ is the adjacency matrix of the graph, with elements ${a_{{\rm{ij}}}} = 1$ means there exists the edge $\left( {{v_i},{v_j}} \right) \in E$; ${\bf{D}} \in {{\mathbb R}^{{\rm{n}} \times {\rm{n}}}}$ is the diagonal degree matrix for ${\bf{A}}$, where ${d_{{\rm{ii}}}} = \sum\nolimits_i {{a_{{\rm{ij}}}}}$, and ${\bf{I}}$ is the identity matrix.
\paragraph{Transformer operating}
The standard structure of Transformer has two key parts, self-attention part and feed-forward network (FFN) part, with $mh\_Attention\left(  \cdot  \right)$ and $FFN\left(  \cdot  \right)$ denote above operations, as defined as follows.
\begin{equation}
mh\_Attention\left( x \right) = Concatenate\left( {Attention\left( x \right)} \right)
\label{eq:attention}
\end{equation}
\begin{equation}
FFN\left( x \right) = max\left( {0,x{W_1} + {b_1}} \right){W_2} + b
\label{eq:ffn}
\end{equation}
\begin{equation}
Attention\left( x \right) = softmax\left( {\frac{{Qx{K^T}x}}{{\sqrt {{d_k}} }}} \right)Vx
\label{eq:mh}
\end{equation}
Where $Q$, $K$ and $V$ denote the weighted matrices, $softmax\left(  \cdot  \right)$ is the softmax operations, ${d_k}$ is the dimension of input $x$, ${W_1}$ and ${W_2}$ are weighted matrices for linear projection, and ${b_1}$ and $b$ denotes biases.
\section{Proposed Method}
\label{sec:method}
\begin{figure*}[h]
    \centering
    \includegraphics[width=0.9\linewidth]{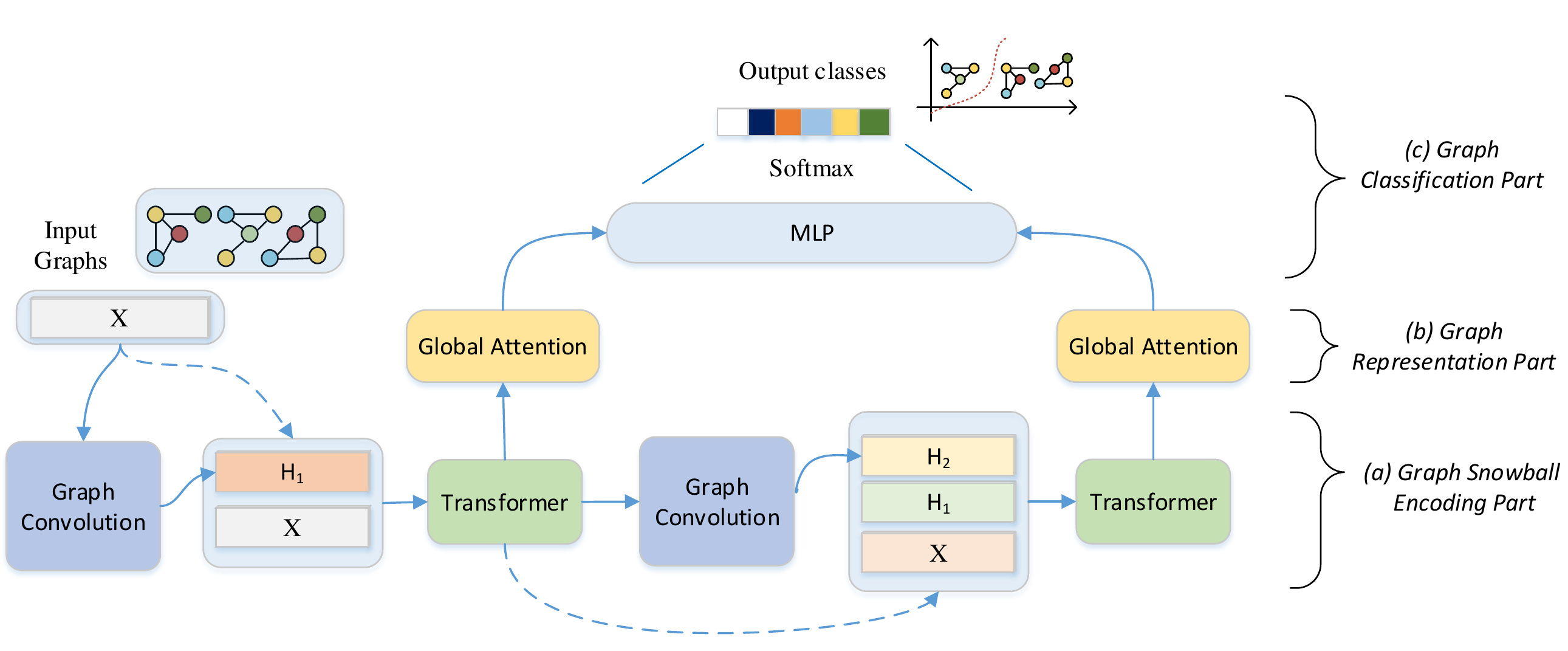}
    \caption{Architecture of TSEN. with three parts, graph snowball encoding part, graph representation part and graph classification part. (a) Graph snowball encoding part has multiple graph snowball encoding layers, and each layer uses graph convolution operation to capture local patterns, and concatenates features from previous layers to obtain multi-scale information, and then uses Transformer to capture global information. (b) Graph representation part uses global attention to read-out the node features from the graph snowball encoding layers, and then concatenates all the attention information. (c) Graph classification part uses a MLP to obtain the probability of classes of the graphs.}
    \label{fig:model}
\end{figure*}

In this paper, we proposed a Transformer and snowball
encoding networks, namely TSEN, with the mixture structure of snowball graph convolution and Transformer. In TSEN, the snowball graph convolution could be regarded as the substitute of position embedding in typical Transformer structure, and the Transformer could be regarded as an encoding layer in snowball structure on the other hand. The TSEN consists of graph snowball encoding part, graph representation part, and graph classification part, as shown in Fig.~\ref{fig:model}.
\subsection{Graph snowball encoding part}
The graph snowball encoding part consists of n layers, and each layer includes a snowball convolution sub-module and a Transformer encoder sub-module. The snowball convolution sub-module firstly concatenates the node features from all previous layers, and uses graph convolution operation to aggregate the concatenated features, and then generates node embedding with local structure information. Afterwards, the Transformer encoder sub-module encodes the node embedding with the Transformer to capture global information, regarding the node embedding from snowball convolution sub-module as the node embedding already encoded with the structural information of the graphs, and forwards the encoded output into the next snowball convolution sub-module.

In the snowball convolution sub-module, a snowball graph convolutional operator was used to obtain the convolutional node embeddings, as defined as follows.
\begin{equation}
{{\bf{H}}^{\left( 0 \right)}} = {\bf{X}}\;
\label{eq:snowballh0}
\end{equation}
\begin{equation}
{{\bf{S}}^{\left( i \right)}} = f\left( {\left[ {{{\bf{H}}^{\left( 0 \right)}},{{\bf{H}}^{\left( 1 \right)}},...,{{\bf{H}}^{\left( i \right)}}} \right]{{\bf{W}}^{\left( i \right)}}} \right)
\label{eq:snowball}
\end{equation}
Where $f\left(  \cdot  \right)$ is the graph convolution operation which was defined in (1), ${{\bf{W}}^{\left( i \right)}}$ is the weighted matrix, and $\left[  \cdot  \right]$ denotes the concatenating operation.

The standard structure of Transformer was used in the encoder sub-module. The Transformer encoding operation was defined as follows.
\begin{equation}
{{\bf{H}}^{'\left( i \right)}} = mh\_Attention\left( {LayerNorm\left( {{{\bf{S}}^{\left( i \right)}}} \right)} \right) + {{\bf{S}}^{\left( i \right)}}
\label{eq:encode1}
\end{equation}
\begin{equation}
{{\bf{H}}^{\left( {i + 1} \right)}} = FFN\left( {LayerNorm\left( {{{\bf{H}}^{'\left( i \right)}}} \right)} \right) + {{\bf{H}}^{'\left( i \right)}}
\label{eq:encode2}
\end{equation}
Where $mh\_Attention\left(  \cdot  \right)$ and $FFN\left(  \cdot  \right)$ of the Transformer were defined in (3) and (4), $LayerNorm( \cdot )$ are layer normalization operations, the convoluted node features ${{\bf{S}}^{\left( i \right)}}$ are the inputs.

With the GSE module, our proposed method TSEN avoid the problems of over-smoothing with stacked GNNs and stacked Transformer~\cite{rampavsek2022recipe}. Moreover, the Transformer structure allows to resolve the expressivity bottlenecks caused by over-squashing~\cite{alon2020bottleneck} by allowing information to spread across the graph via full-connectivity, and does not lose structure information of graphs~\cite{rampavsek2022recipe}.

Layer normalization was used before multi-head self-attention and the FFN blocks to improve the stability of the Transformer part~\cite{RN32}, and dropout and residual operations were adapted in this part.
\subsection{Graph representation part}
In the graph representation part, we used a global attention layer to obtain a graph representation vector for each layer of the graph snowball encoding part. The global attention mechanism of this layer introduced an extra global node connecting to all nodes in the graph, which was similar to the "cls" token to gather global information in NLP tasks~\cite{RN15}, and was an efficient way to represent the graph from node embedding~\cite{RN33}. Then we concatenated all the graph representation vectors as the whole-graph representation~\cite{RN34}, which was used as the input for the downstream classification tasks.

In the global attention layers, multi-layer perception (MLP) was used to map the node features to a score, and the weight of the node was calculated with softmax operation. The global attention operating was defined as a weighted summation with the score and the weight from all the nodes, and the operating of Layer $t$ was defined as follows.
\begin{equation}
{{\bf{h}}_t} = \mathop \sum \limits_{i = 1}^n softmax\left( {{h_{gate}}\left( {{H_i}} \right)} \right) \odot {H_i}
\label{eq:globalattention}
\end{equation}
Where $n$ denotes the number of nodes, ${h_{gate}}\left( \cdot \right)$ is a MLP network, mapping the node features to a score to represent the importance of the node in Layer $t$.
Finally, the whole-graph representation was defined by concatenating the global attention vectors from all the layers, as defined as the follows.
\begin{equation}
{{\bf{h}}_{\cal G}} = \left[ {{{\bf{h}}_1},{{\bf{h}}_2},...{{\bf{h}}_t}} \right],t = 0,1,2,...,n
\label{eq:concat}
\end{equation}
\subsection{Graph classification part}
In the graph classification part, the whole-graph representation was used as the input for MLP and softmax to generate classification result.
\section{Experiments}
\label{sec:experiments}
To evaluate the performance of our proposed TSEN model, we experimentally compared with the recent state-of-the-art models in public brain imaging datasets, and also performed ablation studies. Furthermore, we analyzed the model by evaluating the representation power and the influence of the hyper-parameters.
\subsection{Datasets}
In this paper, we used two human brain functional networks, as shown in Table 1. 

ABIDE I~\cite{RN35} contains the resting-state fMRI data from Autistic Spectrum Disorder (ASD) patients and health controls (HC). By eliminating the low quality of imaging~\cite{RN36}, we selected 974 subjects, including 467 ASD and 507 HC, to construct brain functional networks for classification experiments. 

REST-meta-MDD~\cite{RN37} includes brain network data generated from the resting-state fMRI data from Major Depressive Disorders (MDD) patients and HC. We selected 2027 subjects, including 1041 MDD and 986 HC, to construct brain functional networks for classification experiments. In the datasets of ABIDE I and REST-meta-MDD, we constructed a brain functional network for each subject, where the node represents the ROI based on a brain atlas ~\cite{craddock}, and the edge between nodes represents the brain functional connection of ROI pairs~\cite{kan2022brain, RN39}. The functional connection was defined by the pairwise Pearson correlation coefficient between the mean fMRI series of ROIs. The node/ROI features were defined by the node’s corresponding row in the adjacency matrix of brain network.

We compared TSEN with baselines of three types, including typical GNN models, GNN models with special design for graph classification, and graph-transformer based GNN models.
\begin{itemize}
		\item[$\bullet$] Graph convolutional networks (GCN)~\cite{RN42}, graph attention networks (GAT)~\cite{RN43}, GraphSAGE~\cite{RN44}, and snowball GCN (SBGCN)~\cite{RN18}: There are typical GNN models that were not specifically designed for graph classification. GCN is based on a variant of CNNs which operate directly on graphs. GAT is a GNN architecture using masked self-attentional layers. GraphSAGE is based on randomly edges sampling to enhance the ability of capturing global structure. SBGCN is a GCN architecture using a snowball structure to capture multi-scale information.
		\item[$\bullet$] DiffPool~\cite{RN7}, GIN~\cite{RN34}, PGCN~\cite{RN28}, MRGNN~\cite{RN45}, BrainGNN~\cite{li2021braingnn}, BrainNetCNN~\cite{kawahara2017brainnetcnn}, and FBNetGNN~\cite{kan2022fbnetgen}: There are graph neural networks which have a special design for graph classification. GIN is a well-known typical GNN that was expected to achieve the ability as the Weisfeiler-Lehman graph isomorphism test, but a graph-level readout function of GIN was specially designed to produce the embedding of the entire graph for graph classification tasks. DiffPool is a differentiable graph pooling module that could generate hierarchical representations for graph classification. PGCN introduced the polynomial graph convolution layer to independently exploit neighbouring nodes at different topological distances, and used subsequent readout layers to generate graph representation. MRGNN used a shallow readout function to generate an unsupervised graph representation. BrainGNN was specially designed with graph convolution and interpretation for brain network classification. BrainNetCNN proposed multiple types of convolution to extract features from edge weights. FBNetGNN applied GNNs based on a learnable graph, and the graphs could be considered as a type of attention score.
		\item[$\bullet$] BrainNetTF~\cite{kan2022brain}, GraphGPS~\cite{rampavsek2022recipe}, GraphTrans~\cite{RN15}, and Spectral Attention Network (SAN)~\cite{RN46}: They are well-known graph-transformer based models. GraphTrans applied graph convolution operation to introduce local patterns and structure information of graphs. Graphormer specially designed three positional encoding methods for graph classification. SAN used learned positional encoding of Laplacian spectrum to introduce position information of nodes in graphs. BrainNetTF learned fully pairwise attention weights with Transformer-based models to obtain the brain network structures.
\end{itemize}
\begin{table}[t]
\setlength{\tabcolsep}{3pt}
\centering
\caption{The Characteristic of Graph Datasets}
\label{tab:dataset}
\begin{threeparttable} 
\begin{tabular}{cccccc}
\hline
Dataset       & Classes & Graphs & Avg. Nodes & Avg. Edges & Attr. \\ \hline
ABIDE I       & 2       & 974    & 200.00     & 2362.25    & 200   \\
REST-meta-MDD & 2       & 2027   & 200.00     & 39692.55   & 200   \\
\hline
\end{tabular}
\begin{tablenotes} 
\item “Attr.” means the dimension of the node features. 
\end{tablenotes} 
\end{threeparttable} 
\end{table}
\subsection{Experiment setting}
For the implementation, we used pytorch-geometric~\cite{RN47} as the backend for all typical GNN models, and used pytorch as the backend of transformer structure for all graph-transformer methods. GELU~\cite{RN48} was selected as the activation function of transformer encoder to further improve the unlinearity expressive power compared to ReLU. The batch normalization and dropout operations were added for every fully connected layer in MLP module, and AdamW~\cite{RN49} served as the optimizer. For loss function, we selected cross-entropy for the brain network experiments. During training, linear increasing warm-up and inverse square root decay strategy were selected as the scheduler for the learning rate~\cite{RN51} during the procedure of training.

For the input graphs of the compared models previously designed for brain networks, we adapted their original settings, where the fully connected binarized graphs of brain networks were used for BrainNetTF and FBNetGNN, and the brain network adjacency matrices (i.e., fully connected weighted graphs) were used for BrainGNN and BrainNetCNN. For the input graphs of other compared models, we adapted the same settings as our proposed TSEN, where the edges were defined by the binarization with fixed threshold 0.4 for REST-meta-MDD and 0.5 for ABIDE I respectively. 

For all experiments, the learning rate, weight decay, and batch size were set as 1, 0.0001, and 16, respectively. The dropout rate for fully connected layers in MLP module was set as 0.5, and dropout rate for transformer was set as 0.1. We trained with 300 epochs for the brain network datasets. 

To evaluate the performance of the models, we randomly selected 80\% of the dataset as the training set, 10\% as the validation set, 10\% as the testing set, and ran 5 times to evaluate the performance. Moreover, we used accuracy and F1 score as the metric for the classification models. For the implementation details, we made our source codes publicly available on our project website \href{https://github.com/largeapp/TSEN}{https://github.com/largeapp/TSEN}.
\subsection{Experiment Results}
The results of the experiment are shown in the Table 2. The proposed TSEN achieved the best performance compared with four typical GNN models, seven GNN models with special design for graph classification, and four graph-transformer based GNN models, in both two brain functional network datasets. That indicated the effectiveness of TSEN in capturing the whole-graph representation for brain functional network classification. GraphSAGE, MRGNN, and BrainGNN achieved excel performance, which indicated the GNN methods still worked well on brain networks data. Moreover, BrainNetTF also achieved excel performance compared to other methods, which indicated the power of Transformer on brain network classification. 
\begin{table*}[h]
\label{tab:performance}
\centering
\caption{Accuracy and F1 Score ($\%$) of Graph Classification ($\pm$std)}
\begin{threeparttable} 
\begin{tabular}{ccccccc}
\hline
\multirow{2}{*}{Type}                                      & \multirow{2}{*}{Method}   & \multicolumn{2}{c}{ABIDE I}      & REST-meta-MDD  & \multicolumn{2}{c}{}               \\ \cline{3-7} 
                                                           &                           & ACC            & F1              & ACC            & \multicolumn{2}{c}{F1}             \\ \hline
\multirow{4}{*}{Typical GNNs}  & GCN                       & 65.97$\pm$3.20 & 68.11$\pm$2.12  & 64.73$\pm$2.16 & \multicolumn{2}{c}{61.38$\pm$3.01} \\
                                                           & GAT                       & 67.43$\pm$1.54 & 66.64$\pm$3.01  & 64.53$\pm$2.51 & \multicolumn{2}{c}{64.84$\pm$3.10} \\
                                                           & GraphSAGE                 & 68.33$\pm$2.98 & 71.41$\pm$3.40  & 62.23$\pm$2.17 & \multicolumn{2}{c}{63.36$\pm$0.93} \\
                                                           & SBGCN                     & 64.17$\pm$2.62 & 65.38$\pm$4.10  & 65.19$\pm$2.75 & \multicolumn{2}{c}{65.34$\pm$4.09} \\ \hline
\multirow{7}{*}{GNNs for Graph Classification} & GIN                       & 64.22$\pm$4.32 & 63.31$\pm$12.29 & 56.09$\pm$3.40 & \multicolumn{2}{c}{62.41$\pm$8.38} \\
                                                           & DiffPool                  & 64.48$\pm$3.01 & 65.92$\pm$5.14  & 64.05$\pm$2.68 & \multicolumn{2}{c}{61.84$\pm$2.40} \\
                                                           & PGCN                      & 62.65$\pm$3.12 & 63.30$\pm$5.89  & 62.49$\pm$3.01 & \multicolumn{2}{c}{61.41$\pm$4.23} \\
                                                           & MRGNN                     & 66.11$\pm$2.22 & 67.54$\pm$2.55  & 63.41$\pm$1.24 & \multicolumn{2}{c}{63.94$\pm$4.06} \\ 
                                                            & BrainGNN                 & 67.35$\pm$6.01 & 66.85$\pm$7.60  & 59.80$\pm$2.59 & \multicolumn{2}{c}{53.93$\pm$4.79} \\
                                                           & BrainNetCNN             & 63.16$\pm$3.53 & 64.69$\pm$7.73  & 60.99$\pm$1.44 & \multicolumn{2}{c}{60.98$\pm$3.19} \\
                                                           & FBNetGNN                & 61.05$\pm$1.75 & 61.71$\pm$4.56  & 63.05$\pm$2.16 & \multicolumn{2}{c}{62.84$\pm$1.05} \\ \hline
 \multirow{4}{*}{Transformer based GNNs}           & BrainNetTF              & 65.79$\pm$3.22 & 69.30$\pm$5.04  & 62.76$\pm$1.31 & \multicolumn{2}{c}{65.00$\pm$2.39} \\ 
                                                           & SAN                       & 66.07$\pm$2.93 & 66.60$\pm$3.57  & 61.25$\pm$2.33 & \multicolumn{2}{c}{56.03$\pm$3.75} \\
                                                           & GraphGPS                  & 66.58$\pm$3.85 & 67.20$\pm$3.51  & 65.68$\pm$2.22 & \multicolumn{2}{c}{59.87$\pm$6.36} \\
                                                           & GraphTrans                & 66.92$\pm$2.49 & 64.57$\pm$8.06  & 63.25$\pm$2.54 & \multicolumn{2}{c}{64.91$\pm$3.66} \\ \hline
\multirow{1}{*}{Ours}                               & TSEN                      & \textbf{70.27$\pm$4.34} & \textbf{73.35$\pm$7.00}  & \textbf{66.26$\pm$1.74} & \multicolumn{2}{c}{\textbf{65.75$\pm$1.74}} \\ \hline
\end{tabular}
\begin{tablenotes} 
\item In each row, the highest accuracy and F1 score is highlighted in bold. 
\end{tablenotes} 
\end{threeparttable} 
\end{table*}
\subsection{Ablation Studies}
We performed ablation studies to evaluate the influence of different parts of TSEN model by comparing TSEN with five models, including GCN, SBGCN, SBGCN with FFN part (SBGCN\_FFN), SBGCN with self-attention part (SBGCN\_SA), and GCN with Transformer (GCN\_Trans). GCN\_Trans could be also considered as TSEN without snowball structure. The classification performance of the ablation experiments is shown in Table 3. 
\begin{table}[h]
\label{tab:ablation}
\setlength{\tabcolsep}{3pt}
\centering
\caption{Accuracy ($\%$) and F1 Score ($\%$) of Ablation Study ($\pm$std)}
\begin{threeparttable} 
\begin{tabular}{ccccc}
\hline
\multirow{2}{*}{Method} & \multicolumn{2}{c}{ABIDE I}     & \multicolumn{2}{c}{REST-meta-MDD} \\ \cline{2-5} 
                        & ACC            & F1             & ACC             & F1              \\ \hline
GCN                     & 65.97$\pm$3.20 & 68.11$\pm$2.12 & 64.73$\pm$2.16  & 61.38$\pm$3.01  \\
SBGCN                   & 64.17$\pm$2.62 & 65.38$\pm$4.10 & 65.19$\pm$2.75  & 65.34$\pm$4.09  \\
SBGCN\_FFN              & 69.61$\pm$4.98 & 70.16$\pm$5.31 & 64.89$\pm$3.59  & 65.21$\pm$4.70  \\
SBGCN\_SA               & 68.37$\pm$2.73 & 69.31$\pm$8.06 & 64.53$\pm$0.70  & 65.71$\pm$2.57  \\
GCN\_Trans              & 64.87$\pm$2.77 & 67.10$\pm$2.66 & 63.22$\pm$3.01  & 64.99$\pm$4.97  \\
TSEN                    & 70.27$\pm$4.34 & 73.35$\pm$7.00 & 66.26$\pm$1.74  & 65.75$\pm$1.74  \\ \hline
\end{tabular}
\end{threeparttable} 
\end{table}

Compared with the other five models, the proposed TSEN achieved the best performance. The models with self-attention (e.g., SBGCN\_SA and TSEN) had better performance than the models without self-attention, and GCN with Transformer achieved better performance than GCN or Snowball GCN without Transformer. That indicated the potential of Transformer structure for brain network classification. The GCN with Transformer but without snowball connection was still not better than TSEN which combined Transformer and snowball convolution connection. That indicated the power of the combination of snowball and Transformer in graph classification.
\section{Model Analysis}
\label{sec:analysis}
In this section, we analyzed the representation power of TSEN by T-distributed Stochastic Neighbor Embedding (T-SNE) visualization~\cite{RN52}, and centered kernel alignment (CKA) similarity~\cite{RN53}. And we analyzed the influence of input graphs from brain functional networks for TSEN.
\subsection{The representation power of TSEN}
\begin{figure*}[h]
    \centering
    \includegraphics[width=\linewidth]{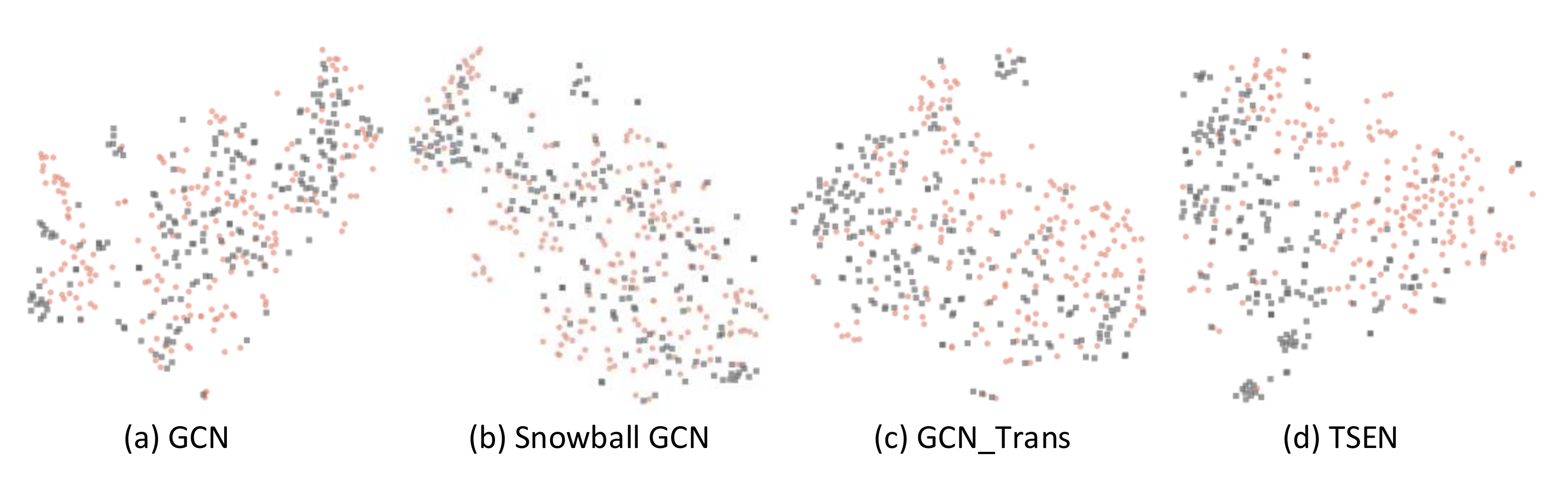}
    \caption{The T-SNE visualization of graph representation. Each point represents a graph, and the colors indicate different classes of graphs.}
    \label{fig:tsne}
\end{figure*}

We compared the representation of four models, including GCN, snowball GCN, GCN with Transformer, and TSEN, using T-SNE visualization in NCI1 dataset. The graph representation was obtained after concatenating all read-out outputs and before inputting into MLP module for classification. The T-SNE visualization of these models are shown in Fig.~\ref{fig:tsne}. 

From Fig.~\ref{fig:tsne}(a), ~\ref{fig:tsne}(b), and ~\ref{fig:tsne}(c), the representations generated by GCN, Snowball GCN, GCN with Transformer were ambiguous and hard to discriminate. From Fig.~\ref{fig:tsne}(d), compared with other models, the representation of TSEN was easier to distinguish, and that indicated the power of representation of TSEN. 

Based on CKA RBF kernel, we calculated the CKA similarity value of representation between different models on the ABIDE I dataset. As shown in Table 4, we compared CKA similarity among models in the first layer and the second layer respectively. The results showed that the CKA similarity between TSEN with other models was relatively smaller than the others. The smaller similarity value indicated that the representation of TSEN was more different than other models, or more powerful representation of the TSEN model.
\begin{table}[h]
\label{tab:difference}
\centering
\caption{CKA Similarity Between Models}
\begin{threeparttable} 
\begin{tabular}{ccc}
\hline
Between Models           & First Layer & Second Layer \\ \hline
SBGCN and SBGCN\_SA      & 0.1251      & 0.0458       \\
SBGCN and SBGCN\_FFN     & 0.0595      & 0.0426       \\
SBGCN\_SA and SBGCN\_FFN & 0.0218      & 0.0356       \\
TSEN and SBGCN           & 0.0867      & 0.0234       \\
TSEN and SBGCN\_SA       & 0.0387      & 0.0267       \\
TSEN and SBGCN\_FFN      & 0.0188      & 0.0135       \\ \hline
\end{tabular}
\end{threeparttable} 
\end{table}

\subsection{The influence of input graphs on TSEN}
We evaluated the influence of different input graphs from brain functional networks on the TSEN model. We binarized the input graphs by applying thresholds on the edges, and used the binarized graphs as input for the graph snowball encoding operation in TSEN. We selected different threshold from 0 to 0.5 to construct binarized graphs of brain networks, and the results are shown in Table 5.
\begin{table}[h]
\label{tab:hyp}
\centering
\caption{Accuracy ($\%$) and F1 Score ($\%$) of TSEN with Different Threshold in GSE operations ($\pm$std)}
\begin{threeparttable} 
\begin{tabular}{ccccc}
\hline
\multirow{2}{*}{Threshold} & \multicolumn{2}{c}{ABIDE I}     & \multicolumn{2}{c}{REST-meta-MDD} \\ \cline{2-5} 
                           & Acc            & F1             & acc             & F1              \\ \hline
0                          & 67.57$\pm$2.46 & 70.98$\pm$3.17 & 55.67$\pm$1.04  & 64.29$\pm$6.82  \\
0.1                        & 66.33$\pm$1.77 & 61.20$\pm$4.04 & 60.10$\pm$2.22  & 65.53$\pm$5.47                \\
0.2                        & 62.89$\pm$2.88        & 66.67$\pm$4.34               &  63.55$\pm$1.39      &   64.49$\pm$3.54               \\
0.3                        & 62.24$\pm$3.30      & 67.83$\pm$5.53               & 62.17$\pm$1.45           & 64.07$\pm$4.80                \\
0.4                        & 68.37$\pm$1.44       & 66.46$\pm$3.07               &  \textbf{66.26$\pm$1.74}  &  \textbf{65.75$\pm$1.74}  \\
0.5                        & \textbf{70.27$\pm$4.34} & 73.35$\pm$7.00 &   62.56$\pm$2.44             & 65.59$\pm$0.26                \\ \hline
\end{tabular}
\begin{tablenotes} 
\item In each row, the highest accuracy and F1 score is highlighted in bold. 
\end{tablenotes} 
\end{threeparttable} 
\end{table}

Compared with other thresholds, the TSEN models achieved the best performance with the threshold of 0.5 for ABIDE I dataset and the threshold of 0.4 for REST-meta-MDD. These threshold did not show a linear relationship with the performance of the model. For the threshold of 0, i.e. without threshold, the model could not achieved better performance compared with some suitable thresholds. That indicated the threshold did have significant influence of the performance of the model; and the graph snowball encoding operation of TSEN could learn valid graph structure information from the brain functional networks, which could be integrated as position embedding for further processing by Transformer neural networks. 

\section{Conclusion}
\label{sec:conclusion}
To learn an effective representation for brain functional network classification, we proposed TSEN, a novel graph Transformer model to capture whole-graph representation with multi-scale information for classification. We evaluated the proposed TSEN model by two large-scale brain functional network datasets, and the results showed that the TSEN model outperformed the state-of-the-art GNN models and the graph-transformer based GNN models. The results demonstrated the combination of graph snowball structure and graph Transformer could enhance the power to capture global patterns of the brain networks. That also demonstrated using graph convolution as position embedding in Transformer structure was a simple yet effective method to capture structure patterns from brain networks naturally.

For future work, we would investigate the connections between the graph properties and the graph Transformer architectures, which can help better understanding the strong potential of graph Transformer models. Besides, we would further evaluate TSEN on more brain network data at voxel or region levels for predicting brain disorders.
\section{Acknowledgements}
\label{sec:ack}
This work was supported in part by the Natural Science Foundation of Guangdong Province under Grant 2021A1515011942, and the Innovation Fund of Introduced High-end Scientific Research Institutions of Zhongshan under Grant 2019AG031.
\section{Declaration of competing interests}
\label{sec:Conflict}
The authors declare that they have no competing interests.
\bibliographystyle{unsrt}  
\bibliography{references}  


\end{CJK}
\end{document}